\documentclass[conference]{IEEEtran}
\IEEEoverridecommandlockouts
\usepackage{cite}
\usepackage{amsmath,amssymb,amsfonts}
\usepackage{algorithmic}
\usepackage{graphicx}
\usepackage{textcomp}
\usepackage{xcolor}
\def\BibTeX{{\rm B\kern-.05em{\sc i\kern-.025em b}\kern-.08em
    T\kern-.1667em\lower.7ex\hbox{E}\kern-.125emX}}
\begin{document}

\title{FireLite: Leveraging Transfer Learning for Efficient Fire Detection in Resource-Constrained Environments\\}

\author{
\IEEEauthorblockN{1\textsuperscript{st} Mahamudul Hasan}
\IEEEauthorblockA{
\textit{Electronic and Electrical Engineering} \\
\textit{University of Chester}\\
Chester, United Kingdom \\
m.h.eee.012@gmail.com}
\and
\IEEEauthorblockN{2\textsuperscript{nd} Md Maruf Al Hossain Prince}
\IEEEauthorblockA{
\textit{Dept. of Computer Science and Engineering}\\
\textit{Islamic University of Technology}\\
Gazipur, Bangladesh \\
marufhossain@iut-dhaka.edu}
\and
\IEEEauthorblockN{3\textsuperscript{rd} Mohammad Samar Ansari}
\IEEEauthorblockA{
\textit{University of Chester}\\
United Kingdom \\
m.ansari@chester.ac.uk}
\and

\IEEEauthorblockN{4\textsuperscript{th} Sabrina Jahan}
\IEEEauthorblockA{
\textit{Dept. of Computer Science and Engineering}\\
\textit{Bangladesh Army University}\\ \textit{of Science and Technology}\\
Saidpur, Bangladesh \\
sabrinasammi06@gmail.com}
\and
\IEEEauthorblockN{5\textsuperscript{th} Abu Saleh Musa Miah}
\IEEEauthorblockA{
\textit{School of Computer Science} \\ \textit{and Engineering}\\
\textit{The University of Aizu}\\
Aizuwakamatsu, Japan \\
abusalehcse.ru@gmail.com}

\and
\IEEEauthorblockN{*6\textsuperscript{th}Jungpil Shin}
\IEEEauthorblockA{
\textit{School of Computer Science} \\ \textit{and Engineering}\\
\textit{The University of Aizu}\\
Aizuwakamatsu, Japan \\
jpshin@u-aizu.ac.jp}
}

\maketitle
\begin{abstract}
Fire hazards are extremely dangerous, particularly in sectors such the transportation industry where political unrest increases the likelihood of their occurring. By employing IP cameras to facilitate the setup of fire detection systems on transport vehicles losses from fire events may be prevented proactively. However, the development of lightweight fire detection models is required due to the computational constraints of the embedded systems within these cameras. We introduce "FireLite," a low-parameter convolutional neural network (CNN) designed for quick fire detection in contexts with limited resources, in answer to this difficulty. With an accuracy of 98.77\%, our model—which has just 34,978 trainable parameters—achieves remarkable performance numbers. It also shows a validation loss of 8.74 and peaks at 98.77 for precision, recall, and F1-score measures. Because of its precision and efficiency, FireLite is a promising.
\end{abstract}


\section{Introduction}
Fire occurrences provide a constant risk to many different industries, but the transportation sector is especially vulnerable to the devastating effects of these events \cite{uugurlu2016analysis}. Politically unstable locations are known to have higher rates of fire dangers, which can result in higher risks and costs for the transportation industry \cite{collins2005households}. Proactive steps must be taken to quickly identify and reduce fire threats since they have the potential to cause accidents on everything from passenger trains to cargo ships. Conventional fire detection techniques, which depend on human interaction or crude alarm systems, frequently fail to provide prompt responses, leading to serious damage and, in certain circumstances, the sad loss of life \cite{youngjin_kim__2021}. However, the integration of intelligent systems into already-existing infrastructure, like IP cameras, presents a particularly promising route for improving fire detection capabilities, given the advances in computer vision and deep learning \cite{foggia2015real}.

Implementing IP cameras that are fitted with fire detection models is a compelling way to improve safety protocols in the transportation industry. Through the utilization of IP cameras' widespread availability and their capacity to get real-time, high-definition footage, automated fire detection systems that can anticipate possible threats may be put into place.

Nevertheless, creating fire detection models appropriate for IP camera embedded systems deployment is a significant problem. These devices usually have limited memory and processing capacity thus it is necessary to build lightweight versions that can function well without sacrificing detection accuracy. This paper makes the following contributions:

\begin{itemize}
    \item We present "FireLite," a low-parameter convolutional neural network (CNN) designed for fast fire detection in contexts with limited resources. For real-time fire detection applications in the transportation industry, FireLite provides a practical solution by refining the model architecture to reduce processing needs while maintaining detection accuracy.
    \item We have utilized the technique of transfer learning, a novel approach that had not been previously integrated into this particular field to the best of our understanding.
    \item We provide complete experimental findings showing FireLite's effectiveness in accurately and efficiently identifying fire occurrences with low computing overhead, as well as in-depth insights into the design and implementation of the system.
\end{itemize}
The remainder of this paper is organized as follows: Section \ref{RW} provides an overview of related work in the field of fire detection and highlights the significance of lightweight models for practical deployment. Section \ref{PA} presents the methodology employed in the design and training of FireLite, including dataset in Subsection \ref{DS} and model architecture in Subsection \ref{MA}. In section \ref{RnD}, we present the evaluation results of FireLite compared to existing approaches. Finally, Section \ref{Con} concludes the paper with a summary of findings and outlines directions for future research in the field of fire detection and prevention.

\section{Related Work} \label{RW}
Many researchers have been using deep learning modules in numerous domains because of their excellence \cite{miah2023dynamic_graph_general,miah2023dynamic_mcsoc,miah2023skeleton_euvip,miah2024sign_largescale,miah2024spatial_paa,mallik2024virtual_rahim_miah,rahim2024advanced_miah,electronics12132841_miah_multistream_4}. Lightweight fire detection models for real-time applications have advanced significantly in the last few years, especially in the context of Internet of Things (IoT) devices. The subsequent segment offers a thorough synopsis of noteworthy advancements in this field, emphasizing pivotal techniques, datasets employed, and performance indicators attained by every strategy.

FireNet is a lightweight fire and smoke detection model that was created especially for real-time Internet of Things applications by Jadon et al. \cite{jadon2019firenet} by utilizing the FireNet dataset \cite{jadon2019firenet}, which consists of 16 videos without fire and 46 videos with fire, the model was able to reach a parameter count of 646,818. The model exhibited remarkable performance measures, including 96.53\% accuracy, 97\% precision, 94\% recall, and 95\% F1-score. On the other hand, the false negative rate was recorded at 4.13\%, and the false positive rate at 1.95\%. Shees et al. \cite{shees2023firenet} suggested FireNet-v2, an improved lightweight fire detection model designed for real-time IoT applications, building on the framework established by FireNet. Using 363 fire and 3201 non-fire images from the FireNet and Foggia datasets \cite{foggia2015real} combined, the model reached a parameter count of 318,460. Impressively, FireNet-v2 \cite{shees2023firenet} demonstrated notable gains in performance, with an accuracy of 98.43\%. However, they did not use additional matrices for evaluating performance, such as precision, recall, F-1 score, false positive, and false negative. This may cause a model to become overfit. FireNet-Tiny, a very effective fire detection model with a very tiny parameter count, was suggested by Oyebanji et al. \cite{oyebanji2023firenet} By utilizing the FireNet dataset, the model was able to reach a count of 261,922 parameters. With an accuracy of 95.75\%, FireNet-Tiny showed remarkable performance in spite of its small design. Even while current techniques have advanced the creation of lightweight fire detection models significantly, each has its own advantages and disadvantages. Interestingly, both models have different counts of parameters—646,818 in FireNet and 171,234 in FireNet-Micro, for example—which emphasizes how crucial model efficiency is. Specifically, FireLite, our suggested model, stands out with 34,978 parameters count and comparable performance metrics. FireNet- Micro \cite{marakkaparambil2023firenet}, which is equipped with a total of 171,234 parameters, demonstrates superior performance in terms of accuracy when compared to alternative models. This is evidenced by its remarkable achievement of 96.78\% accuracy coupled with an impressive recall rate of 98\%. Nevertheless, it is important to note that FireNet-Micro displays a lower precision level, standing at 89.90\%, along with an F1-score of 93.77\%.

By means of extensive testing and assessment, we prove that FireLite is effective in tackling the computational limitations present in embedded systems, providing a workable solution for quick fire detection in contexts with limited resources.

\section{Proposed Approach} \label{PA}
Using the FireNet Dataset \cite{jadon2019firenet}, we developed a lightweight custom CNN model with a smaller trainable parameter and high performance. More information about the dataset we have used and our CNN model is provided below

\subsection{Dataset} \label{DS}
We utilized the FireNet dataset for the training and testing of our model. The description the dataset is provided below:

\subsubsection{FireNet Dataset}
\begin{figure}[htbp]
\centerline{\includegraphics[width=0.9\linewidth]{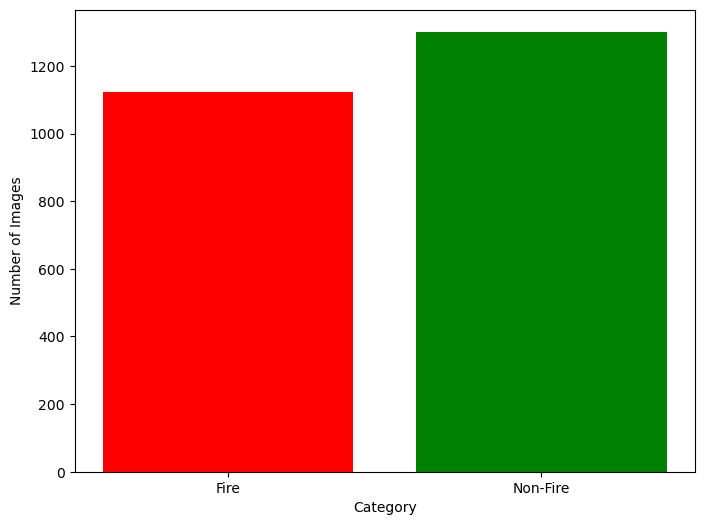}}
\caption{Number of Fire and Non-Fire Images.}
\label{image count}
\end{figure}

\begin{figure}[htbp]
\centerline{\includegraphics[width=0.9\linewidth]{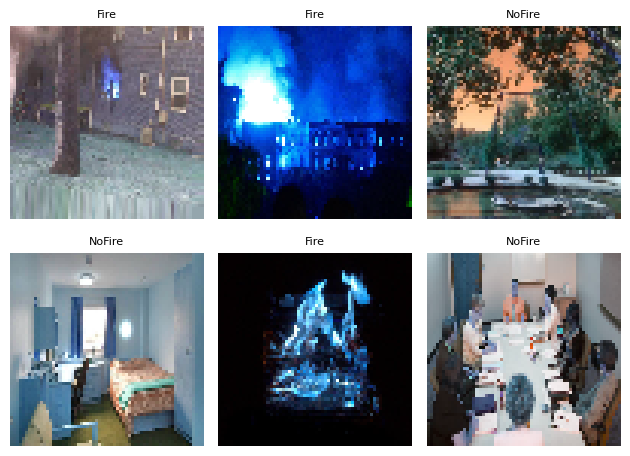}}
\caption{Number of Fire and Non-Fire Images.}
\label{sample}
\end{figure}

We have employed this dataset for the purposes of training, validation and testing. The entire dataset comprises of 16 non-fire films (6,747 frames), and 46 fire videos (19,094 frames). There are 1,124 images of fire and 1,301 images of non-fire in the collection. Fig.\ref{image count}  shows a bar chart of the number of samples. Despite the dataset's seeming tiny size, it is incredibly diversified. The images in the dataset are a combination of photos from the internet (Flickr and Google) plus a small sample of fire and non-fire images from the datasets of Foggia et al. \cite{foggia2015real} and Sharma et al. \cite{sharma2017deep} To preserve the dataset's diversity, Sharma's dataset was augmented and random selected a few images. Fig. \ref{sample} depicts a few images from the sample dataset.

\begin{figure}[tbp]
\centerline{\includegraphics[width=0.7\linewidth]{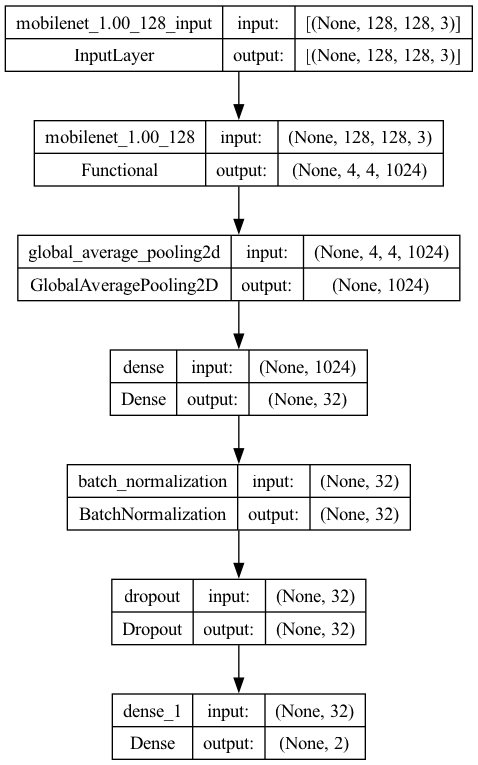}}
\caption{Architecture of the Proposed Model.}
\label{architecture}
\end{figure}

\subsection{Model Architecture} \label{MA}
Fig. \ref{architecture} illustrates the proposed network's whole architectural diagram. We utilized a transfer learning \cite{miah2022bensignnet,hassan2024deep_har_miah,10624624_lstm_najmul_miah_har_conference,anom} strategy employing the MobileNet architecture \cite{sinha2019thin} which had been pre-trained on the ImageNet dataset \cite{deng2009imagenet}. The MobileNet model, pre-trained, was employed without its classification layers, functioning as a feature extractor to capture meaningful representations from input images. This approach facilitated the utilization of high-level features that had been learned from a variety of objects in the ImageNet dataset. To customize the model for our particular task, we trained only the top two layers of the MobileNet architecture while keeping the other layers fixed. This strategy of fine-tuning enabled us to adjust the pre-trained features to our dataset while maintaining the generalization abilities acquired from the ImageNet dataset. The initial layer was the MobileNet base model, followed by a GlobalAveragePooling2D layer to decrease the spatial dimensions of the feature maps extracted by the MobileNet backbone. Following this, a Dense layer with 32 units and ReLU activation function was added, enhanced with batch normalization for training stability and dropout regularization (dropout rate = 0.5) to address overfitting. These regularization methods were implemented to enhance the model's resilience and prevent it from memorizing noise in the training data. The output layer consisted of a Dense layer with softmax activation, enabling the prediction of a probability distribution over the target classes for multi-class classification. In our binary classification task, the output layer had two units corresponding to the two target classes. We applied the sparse categorical cross-entropy loss function, appropriate for tasks involving multi- class classification with integer labels. The Adam optimizer was employed for model training, providing adaptive learning rates and momentum optimization to accelerate convergence. Throughout the training process, we evaluated the accuracy metric to gauge the model's performance on both the training and validation sets. Training was conducted across numerous epochs, with a batch size of 32 samples per iteration to effectively utilize computational resources.

In conclusion, our model architecture utilized transfer learning with the MobileNet backbone, fine-tuning specific layers to tailor the pre-trained features to our target task. The integration of regularization techniques aimed to enhance model generalization and combat overfitting, ultimately enabling successful binary classification on our dataset.

\begin{figure}[htbp]
\centerline{\includegraphics[width=0.7\linewidth]{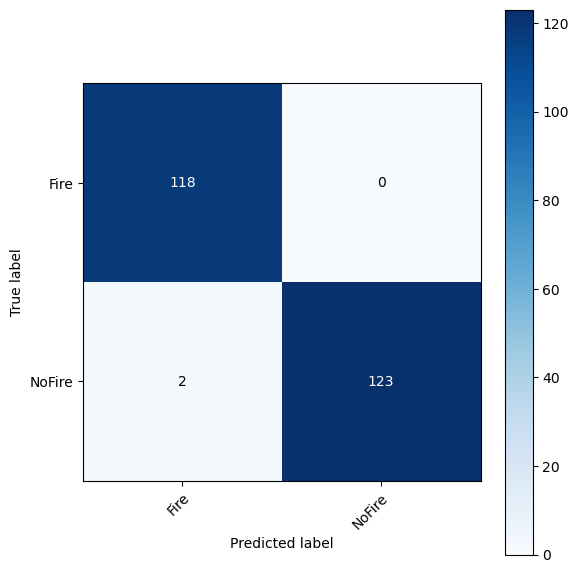}}
\caption{Number of Fire and Non-Fire Images.}
\label{conf_mat}
\end{figure}

\begin{figure}[htbp]
\centerline{\includegraphics[width=0.7\linewidth]{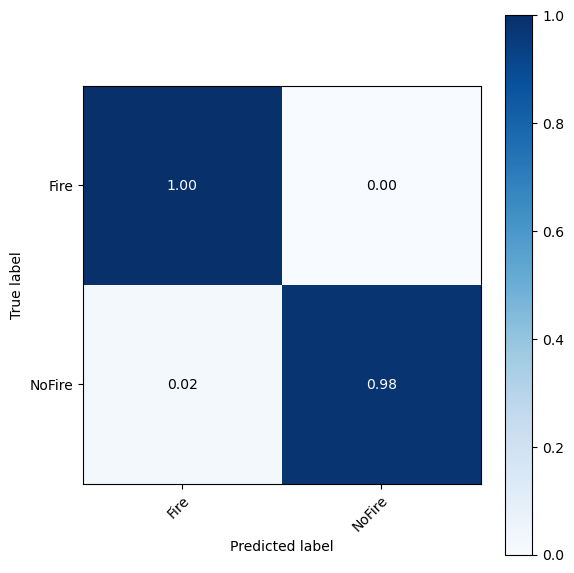}}
\caption{Number of Fire and Non-Fire Images.}
\label{norm_conf}
\end{figure}

\section{Results and Discussion} \label{RnD}
In this examination, the efficacy of the FireNet model was analyzed on the FireNet Dataset. The evaluation criteria addressed a variety of crucial measures such as True Positives (TP), True Negatives (TN), False Positives (FP), False Negatives (FN), Accuracy, Precision, Recall, and F1-Score.

On the FireNet Dataset, the performance of our model was deemed commendable as it achieved an Accuracy of 99.18\%. Upon examination of the confusion matrix, it was observed that there were 118 instances of True Positives and 123 instances of True Negatives, thus indicating a notable level of accurate predictions. Furthermore, the presence of 0 False Negative and 2 False Positives suggested a minimal occurrence of misclassification errors. The Accuracy, Recall, and F1-score were all computed at 99.18\% and Precision of 99.19\%, emphasizing the model's strength in precisely identifying occurrences of fire. Fig. \ref{conf_mat} and Fig. \ref{norm_conf} show the confusion matrix and the normalized confusion matrix, respectively.

With a relatively modest parameter count of 34,978, FireLite demonstrates competitive performance in fire detection. In contrast, FireNet has a substantially greater parameter count of 646,818 and achieves lower accuracy of 93.91\%. Using 318,460 parameters, FireNet v2 obtains an accuracy of 94.95\%. With 171,234 parameters, FireNet-Micro outperforms other models in accuracy, attaining 96.78\%. With an accuracy of 95.75\%, FireNet-Tiny performed well with 261,922 parameters. A thorough comparison is provided in Table \ref{tab1}.

\begin{figure}[tbp]
\centerline{\includegraphics[width=1.0\linewidth]{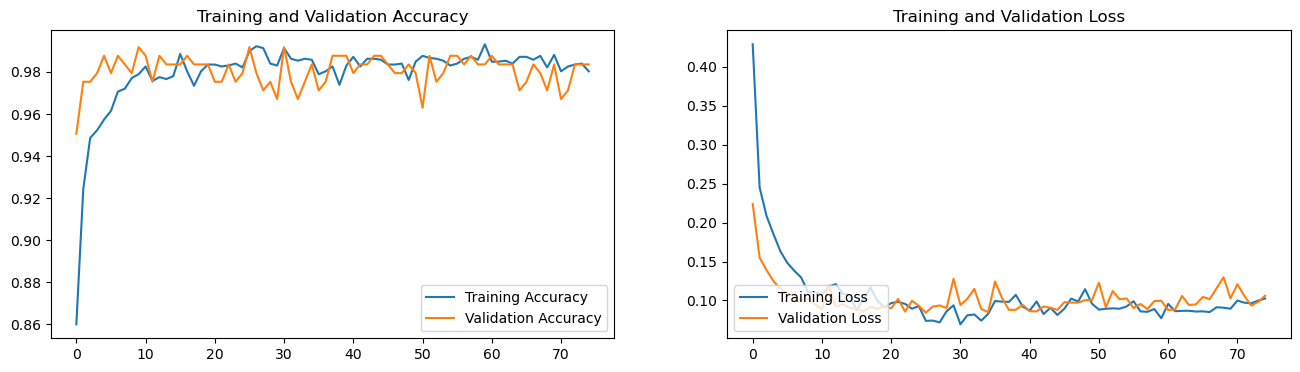}}
\caption{Training vs validation accuracy and loss.}
\label{train_vs_val}
\end{figure}

In addition to assessing the model's performance using metrics and confusion matrices, we also depict the training and validation accuracy, as well as the training and validation loss graphs in Fig. \ref{train_vs_val}, which offer insights into the model's learning progress across epochs, indicating its generalization and error minimization capabilities during training, enabling the identification of overfitting or underfitting issues and facilitating informed decisions on model design and training approaches.

\begin{table}[b]
\caption{Performance comparison of FireLite with other state-of-the art models}
\begin{center}
\begin{tabular}{|c|c|c|c|c|c|c|}
\hline 
\textbf{Model} & \textbf{Accuracy} & \textbf{Parameters} \\
\hline
\textbf{FireLite} & \textbf{99.18} & \textbf{34,978} \\
\hline
FireNet \cite{jadon2019firenet} & 93.91 & 646,818 \\
\hline
FireNet v2 \cite{shees2023firenet} & 94.95 & 318,460 \\
\hline
FireNet Micro \cite{marakkaparambil2023firenet} & 96.78 & 171,234 \\
\hline
FireNet-Tiny \cite{oyebanji2023firenet} & 95.75 & 261,922 \\
\hline
\end{tabular}
\label{tab1}
\end{center}
\end{table}

\section{Conclusion} \label{Con}
In this investigation, we have tackled the crucial issue of fire detection in environments with high risks, especially within the transportation sector, where the presence of fire hazards is exacerbated by political instability. Acknowledging the significance of proactive measures for fire prevention, we have introduced and formulated "FireLite," a lightweight convolutional neural network (CNN) designed for effective fire detection in settings with limited resources. Through the utilization of transfer learning utilizing the MobileNet architecture, which was pretrained on the ImageNet dataset, FireLite has displayed exceptional performance despite its minimal number of parameters. By adjusting specific layers of the MobileNet backbone and integrating regularization methods, FireLite has attained a validation accuracy of 99.18\%, highlighting its accuracy and efficiency in fire detection assignments. Our assessment of FireLite has demonstrated its effectiveness in precisely identifying both fire and non-fire incidents, achieving a high overall accuracy rate. Despite a small number of misclassifications, the performance of FireLite emphasizes its potential for practical implementation in fire detection systems. A comparison with existing models has showcased the competitive performance of FireLite with significantly fewer parameters, positioning it as a promising solution for situations with computational limitations. While other models have displayed varying trade-offs in terms of accuracy, precision, recall, and F1-score, FireLite has stood out due to its balanced performance and efficiency. Further enhancements and enlargement of the training dataset could improve the reliability and effectiveness of FireLite in reducing false alarms and missed detections. Moreover, continual research endeavors can investigate methods to address the identified deficiencies and enhance the robustness of FireLite in various operational conditions.

In conclusion, FireLite marks a notable progression in lightweight fire detection models, providing a dependable and effective resolution for mitigating fire risks in environments with restricted computational capacities. The successful establishment of FireLite underscores the potential of deep learning methodologies in tackling urgent issues in fire safety, paving the way for improved prevention and response tactics in high-risk settings.

\bibliographystyle{IEEEtran}
\bibliography{references}

\vspace{12pt}
\color{red}

\end{document}